\pdfoutput=1

\documentclass[11pt]{article}

\usepackage{acl}
\usepackage{color, soul}
\usepackage{times}
\usepackage{latexsym}

\usepackage[T1]{fontenc}

\usepackage[utf8]{inputenc}

\usepackage{microtype}

\usepackage{multirow}
\usepackage{latexsym}
\usepackage{booktabs}
\usepackage{courier}
\usepackage{amsmath}
\usepackage{subfig}
\usepackage{placeins}
\usepackage{tikz}
\usepackage{hyperref}
\usepackage{amsmath}
\usepackage{amssymb}
\usepackage{bbm}
\usepackage[ruled,vlined]{algorithm2e}

\makeatletter
\newcommand{\printfnsymbol}[1]{%
  \textsuperscript{\@fnsymbol{#1}}%
}
\makeatother

\title{Reasoning with Natural Language Explanations}

\author{Marco Valentino$^{1}$, Andr\'e Freitas$^{1,2,3}$\\  $^{1}$Idiap Research Institute, Switzerland,\\
$^{2}$Department of Computer Science, University of Manchester, UK\\ 
$^{3}$National Biomarker Centre, CRUK-MI, University of Manchester, UK\\
{\tt first.last@idiap.ch} \\}

\begin{document}
\maketitle

\begin{abstract}
Explanation constitutes an archetypal feature of human rationality, underpinning learning and generalisation, and representing one of the media supporting scientific discovery and communication. Due to the importance of explanations in human reasoning, an increasing amount of research in Natural Language Inference (NLI) has started reconsidering the role that explanations play in learning and inference, attempting to build explanation-based NLI models that can effectively encode and use natural language explanations on downstream tasks. 
Research in explanation-based NLI, however, presents specific challenges and opportunities, as explanatory reasoning reflects aspects of both material and formal inference, making it a particularly rich setting to model and deliver complex reasoning.
In this tutorial, we provide a comprehensive introduction to the field of explanation-based NLI, grounding this discussion on the epistemological-linguistic foundations of explanations, systematically describing the main architectural trends and evaluation methodologies that can be used to build systems capable of explanatory reasoning\footnote{Tutorial website: \url{https://sites.google.com/view/reasoning-with-explanations}}.
\end{abstract}

 \section{Introduction}

Building systems that can understand and explain the world is a long-standing goal for \emph{Artificial Intelligence (AI)} \cite{miller2019explanation,mitchell1986explanation,thagard2008models}. The ability to explain, in fact, constitutes an archetypal feature of human rationality, underpinning communication, learning, and generalisation, as well as one of the mediums enabling scientific discovery and progress through the formulation of explanatory theories \cite{lombrozo2012explanation,salmon2006four,kitcher1989explanatory,deutsch2011beginning}.


Due to the importance of explanation in human reasoning, an increasing amount of work has started reconsidering the role that explanation plays in learning and inference with natural language \cite{camburu2018snli,yang2018hotpotqa,rajani2019explain,jansen2018worldtree}. In contrast to the existing end-to-end paradigm based on Deep Learning, explanation-based NLI focuses on developing and evaluating models that can address downstream tasks through the explicit construction of a \emph{natural language explanation} \cite{dalvi2021explaining,jansen2016s,wiegreffe2021teach,stacey2021supervising}. In this context, explanation is seen as a potential solution to mitigate some of the well-known limitations in neural-based NLI architectures \cite{thayaparan2020survey}, including the susceptibility to learning via shortcuts, the inability to generalise out-of-distribution, and the lack of interpretability \cite{guidotti2018survey,biran2017explanation,geirhos2020shortcut,lewis2021question,sinha2021unnatural,schlegel2020framework}.

Research in explanation-based NLI, however, presents several fundamental challenges \cite{valentino2024nature}. 
First, the applied methodologies are still poorly informed by theories and accounts of explanations \cite{salmon2006four,woodwardscientific}. This gap between theory and practice poses the risk of slowing down progress, missing the opportunity to formulate clearer hypotheses on the inferential properties of natural language explanations and define systematic evaluation methodologies \cite{camburu-etal-2020-make,jansen2021challenges,atanasova2024diagnostic}. 
Second, explanation-based NLI models still lack robustness, control, and scalability for real-world applications.
In particular, existing approaches suffer from several limitations when composing explanatory reasoning chains and performing abstraction for NLI in complex domains \cite{khashabi2019capabilities,valentino2022hybrid}.

In this tutorial, we will provide a comprehensive introduction to explanatory reasoning in the context of NLI, by systematically categorising and surveying explanation-supporting benchmarks, architectures, and research trends.
Specifically, we will present how the understanding of explanatory inference have evolved in recent years, together with the emerging methodological and modelling strategies. In parallel, we will attempt to provide an epistemological-linguistic characterisation of natural language explanations reviewing the main theoretical accounts \cite{valentino2024nature,salmon2006four} to derive a fresh perspective for future work in the field. 

\section{Description}
 
This section outlines the content of the tutorial.

\subsection{Epistemological-Linguistic Foundations}

One of the main objectives of the tutorial is to provide a theoretically grounded foundation for explanation-based NLI, investigating the notion of explanation as a language and inference scientific object of interest, from both an \emph{epistemological} and \emph{linguistic} perspectives \cite{valentino2024nature,salmon2006four,jansen2016s}. 

To this end, we will present a systematic survey of the contemporary discussion in Philosophy of Science around the notion of a scientific explanation, attempting to shed light on the nature and function of explanatory arguments and their constituting elements. Here, we will critically review the main accounts of explanations, including the deductive-nomological and inductive-statistical account \cite{hempel1948studies}, the notion of statistical relevance and the causal-mechanical model \cite{salmon1984scientific}, and the unificationist account \cite{kitcher1989explanatory}, aiming to elicit what it means to perform explanatory reasoning.
Following the survey, we will focus on grounding the theoretical accounts for explanation-based NLI, attempting to identify the main feature of explanatory arguments in existing corpora of natural language explanations \cite{jansen2016s,xie2020worldtree,jansen2018worldtree}.

\subsection{Resources \& Evaluation Methods for Explanation-Based NLI}

In order to build NLI models that can reason through the generation of natural language explanations it is necessary to develop systematic evaluation methodologies. To this end, The tutorial will review the main resources, benchmarks and metrics in the field \cite{wiegreffe1teach}. 

Depending on the nature of the NLI problem, an explanation can include pieces of evidence at different levels of abstraction \cite{thayaparan2020survey}. Traditionally, the field has been divided into \emph{extractive} and \emph{abstractive} tasks. In extractive NLI, the reasoning required for the explanations is derivable from the original problem formulation, where the correct decomposition of the problem contains all the necessary inference steps for the answer \cite{yang2018hotpotqa}. On the other hand, abstractive NLI tasks require going beyond the surface form of the problem, where an explanation needs to account for and cohere definitions, abstract relations, which are not immediately available from the original context  \cite{jansen2021challenges,thayaparan-etal-2021-textgraphs}.

In addition, the tutorial will review the main evaluation metrics adopted to assess the quality of natural language explanations. Evaluating the quality of explanations, in fact, is a challenging problem as it requires accounting for multiple concurrent properties. Different metrics have been proposed in the field, ranging from reference-based metrics designed to assess the alignment between automatically generated explanations and human-annotated explanations \cite{camburu2018snli,jansen2021challenges}, and reference-free metrics designed to evaluate additional dimensions such as faithfulness \cite{parcalabescu2024measuring,atanasova2023faithfulness}, robustness \cite{camburu-etal-2020-make}, logical validity \cite{quan2024verification,valentino2021natural}, and plausibility \cite{dalal-etal-2024-inference}. 

\subsection{Explanation-Based Learning \& Inference}

We review the key architectural patterns and modelling strategies for reasoning and learning over natural language explanations. In particular, we focus on the following paradigms:

\paragraph{Multi-Hop Reasoning \& Retrieval-Based Models.} The construction of explanations typically requires multi-hop reasoning -- i.e., the ability to compose multiple pieces of evidence to support the final answer \cite{dalvi2021explaining,xie2020worldtree}. Multi-hop reasoning has been largely studied in a retrieval settings, where, given an external knowledge base, the model is required to select, collect and link the relevant knowledge required to arrive at a final answer \cite{valentino2022hybrid,valentino2021unification,valentino2022case}. Here, we will review the main retrieval-based architectures for multi-hop reasoning and explanation, highlighting some of the inherent limitations of such paradigm, including the tension between semantic drift and efficiency \cite{khashabi2019capabilities}.

\paragraph{Natural Language Explanation Generation.} In parallel with retrieval approaches, NLI using generative models have been used for supporting explanatory inference \cite{camburu2018snli,rajani2019explain}. In this setting, early approaches leverage human-annotated natural language explanations for training generative models \cite{dalvi2021explaining}. Subsequently, the advent of Large Language Models (LLMs) has made it possible to elicit explanatory reasoning via specific prompting techniques and in-context learning \cite{wei2022chain,yao2024tree,zheng2023take,he-etal-2024-using}. Here, we review the main trends in the LLM-based generative paradigms, highlighting persisting limitations such as hallucinations and faithfulness \cite{turpin2024language}.

\subsection{Semantic Control for Explanatory Reasoning}

Controlling the explanation generation process in neural-based models is particularly critical while modelling complex reasoning tasks. In this tutorial, we will review emerging trends which combine neural and symbolic approaches to improve semantic control in the explanatory reasoning process, which can provide formal guarantees on the quality of the explanations. These methods aim to integrate the content flexibility of language models (instrumental for supporting material inferences) and a formal inference properties.

In particular, we focus on the following key methods:

\paragraph{Leveraging Explanatory Inference Patterns for Explanation-Based NLI.} Inference patterns in explanation corpora can be leveraged to improve the efficiency and robustness of neural representations \cite{valentino2024nature,zhang2023towards}. In particular, we will review approaches that attempt to leverage the notion of unification power in corpora of natural language explanations to improve multi-hop reasoning in a retrieval setting and alleviate semantic drift \cite{valentino2022hybrid,valentino2021unification,valentino2022case}.

\paragraph{Constraint-Based Optimisation for Explanation-Based NLI.} We will focus on describing neuro-symbolic methods which target encoding explicit assumptions about the structure of natural language explanations \cite{thayaparan2021explainable}. Here, we will review methods performing multi-hop inference via constrained optimisation, integrating neural representations with explicit constraints via end-to-end differentiable optimisation approaches \cite{thayaparan2022diff,thayaparan2024differentiable}.

\paragraph{Formal-Geometric Inference Controls over Latent Spaces.} Covers emerging methodologies which focus on learning latent spaces with better representational properties for explanatory NLI, using language Variational Autoencoders (VAEs) for delivering better disentanglement and separability of language and inference properties \cite{zhang-etal-2024-learning,zhang-etal-2024-graph,zhang-etal-2024-improving,zhang-etal-2024-learning} which support better inference control. These methods deliver an additional geometrical structure to latent spaces, aiming to deliver the vision of 'inference as latent geometry'.

\paragraph{LLM-Symbolic Architectures} Finally, we will focus on hybrid neuro-symbolic architectures that attempt to leverage the material/content-based inference properties of LLMs for explanation generation with external symbolic approaches, which accounts for formal/logical validity refinement properties. In particular, we will review approaches that perform explanation refinement via the integration of LLMs and Theorem Provers to verify logical validity \cite{quan2024verification,quan-etal-2024-enhancing} and additional external tools to evaluate explanation properties such as uncertainty, plausibility and coherence \cite{dalal-etal-2024-inference}.

\section{Schedule}

The tutorial will be organised according to the following timeline:

\begin{enumerate}
    \item Introduction \& Motivation (20 min.)
    \item Epistemological-Linguistic Foundations (20 min.)
    \item Resources \& Evaluation for Explanation-Based NLI (40 min.)
    \item Explanation-Based Learning \& Inference (40 min.)
    \item Semantic Control for Explanatory Reasoning (40 min.)
    \item Synthesis, Discussion, and Q\&A (20 min.)
\end{enumerate}

 \section{Breadth \& Diversity}
 

The tutorial will cover a wide spectrum of topics in different fields, ranging from Philosophy, Machine Learning, Natural Language Processing, Knowledge Representation and Automated Reasoning. This diversity of topics will help create a rich environment in which academics from different backgrounds and cultural contexts can integrate different perspectives. The tutorial plan includes integrated open Q\&A sessions and practical demonstrations.
  
 \section{Prerequisites}
 
We do not expect attendees to be familiar with previous research on NLI and Explanatory inference. On the opposite, we intent this tutorial to be an efficient and deep onboarding into the state-of-the-art in those areas. Participants should have a general background knowledge in deep learning, including recent trends and architectures such as Large Language Models. Participants are expected to be familiar with some of the broader NLI tasks, such as Textual Entailment and Question Answering.
 
 \section{Reading List}

 \paragraph{Epistemological-Linguistic Foundations}

\paragraph{\citet{valentino2024nature}} On the Nature of Explanation: An Epistemological-Linguistic Perspective for Explanation-Based Natural Language Inference.

\paragraph{\citet{salmon2006four}} Four Decades of Scientific Explanation.

\paragraph{\citet{jansen2016s}} What’s in an Explanation? Characterizing Knowledge and Inference Requirements for Elementary Science Exam.

\paragraph{Resources, Models and Evaluation}
\paragraph{\citet{wiegreffe2021teach}} Teach me to Explain: A Review of Datasets for Explainable Natural Language Processing.
\paragraph{\citet{thayaparan2020survey}} A Survey on Explainability in Machine Reading Comprehension.
\paragraph{\citet{zhao2024explainability}} Explainability for Large Language Models: A Survey.

\paragraph{Related Tutorials}
\paragraph{\citet{zhu-etal-2024-explanation}} Explanation in the Era of Large Language Models.
\paragraph{\citet{camburu2021natural}} Natural-XAI: Explainable AI with Natural Language Explanation.
\paragraph{\citet{zhao-etal-2023-complex}} Complex Reasoning in Natural Language.
\paragraph{\citet{boyd-graber-etal-2022-human}} Human-Centered Evaluation of Explanations.

 \section{Instructor information}
 \paragraph{Marco Valentino, {\normalfont{Idiap Research Institute}}.\protect\footnote{\url{mailto:marco.valentino@idiap.ch}}} Marco is a postdoctoral researcher at the Idiap Research Institute, Switzerland. His research is carried out at the intersection of Natural Language Inference and Neuro-Symbolic models focusing on building systems that can reason through natural language explanations in complex domains (e.g., mathematics, science, biomedical and clinical applications, ethical reasoning). He has published papers in major AI and NLP conferences including AAAI, ACL, EMNLP, NAACL and EACL. Marco was involved in the organisation of workshops including MathNLP (EMNLP 2022 and LREC-COLING 2024), and TextGraphs (COLING 2022 and ACL 2024).

 \paragraph{Andr\'e Freitas, {\normalfont{University of Manchester \& Idiap Research Institute}}.\protect\footnote{\url{mailto:andre.freitas@manchester.ac.uk}}} André Freitas leads the Neuro-symbolic AI Lab at the University of Manchester and IDIAP Research Institute. His main research interests are on enabling the development of AI methods to support abstract, flexible and controlled reasoning in order to support AI-augmented scientific discovery. In particular, he investigates how the combination of neural and symbolic data representation paradigms can deliver better models of inference. He is an active contributor to the main conferences and journals in the AI/Natural Language Processing (NLP) interface (AAAI, NeurIPs, ACL, EMNLP, EACL, COLING, TACL, Computational Linguistics), with over 100 peer-reviewed publications. He contributed to the organisation of MathNLP at EMNLP 2022 and LREC-COLING 2024. Andr\'e participated in 7 tutorials, and co-organised 1 conference and 6 workshops.

\section*{Acknowledgements}
This work was partially funded by the Swiss National Science Foundation (SNSF) project NeuMath (\href{https://data.snf.ch/grants/grant/204617}{200021\_204617}).

\bibliography{acl}
\bibliographystyle{acl_natbib}

\end{document}